\documentclass[letterpaper, 10 pt, conference]{ieeeconf}  

\IEEEoverridecommandlockouts                              

\usepackage{xcolor}
\usepackage{url}
\usepackage{booktabs}
\usepackage{tabularx}
\usepackage{array}
\newcolumntype{Y}{>{\centering\arraybackslash}X} 

\title{\LARGE \bf
Robust Sim-to-Real Cloth Untangling through Reduced-Resolution Observations via Adaptive Force-Difference Quantization
}

\author{Yoshihisa Tsurumine$^{1}$, Yuki Kadokawa$^{1}$, Kohei Hayashi$^{1}$, Christian Diehm$^{1,2}$, Takamitsu Matsubara$^{1}$
\thanks{
    $^{1}$ All the authors are with the Nara Institute of Science and Technology, Japan. $^{2}$ Christian Diehm is with the Karlsruhe Institute of Technology, Germany.
}%
}

\usepackage{setspace}
\usepackage{comment}
\usepackage{subfigure}
\usepackage[font=footnotesize]{caption}
\usepackage{booktabs}
\usepackage{graphicx}
\usepackage{amsmath}
\usepackage{amsfonts}
\usepackage{amssymb}
\usepackage{multirow} 
\usepackage{algorithmic}
\usepackage{bm}
\usepackage[space, compress]{cite}

\usepackage[ruled,vlined]{algorithm2e}
\usepackage{algorithm2e,setspace}

\usepackage[colorlinks=true, linkcolor=blue, citecolor=blue, urlcolor=blue]{hyperref}

\begin{document}

\maketitle
\thispagestyle{empty}
\pagestyle{empty}

\begin{abstract}
Robotic cloth untangling requires progressively disentangling fabric by adapting pulling actions to changing contact and tension conditions. Because large-scale real-world training is impractical due to cloth damage and hardware wear, sim-to-real policy transfer is a promising solution. However, cloth manipulation is highly sensitive to interaction dynamics, and policies that depend on precise force magnitudes often fail after transfer because similar force responses cannot be reproduced due to the reality gap. We observe that untangling is largely characterized by qualitative tension transitions rather than exact force values. This indicates that directly minimizing the sim-to-real gap in raw force measurements does not necessarily align with the task structure. We therefore hypothesize that emphasizing coarse force-change patterns while suppressing fine environment-dependent variations can improve robustness of sim-to-real transfer. Based on this insight, we propose Adaptive Force-Difference Quantization (ADQ), which reduces observation resolution by representing force inputs as discretized temporal differences and learning state-dependent quantization thresholds adaptively. This representation mitigates overfitting to environment-specific force characteristics and facilitates direct sim-to-real transfer. Experiments in both simulation and real-world cloth untangling demonstrate that ADQ achieves higher success rates and exhibits greater robustness in sim-to-real transfer than policies using raw force inputs.
Supplementary video is available at \url{https://youtu.be/ZeoBs-t0AWc}
\end{abstract}


\section{Introduction}

Robotic cloth untangling is a fundamental task for household service robots \cite{LonghiniArXiv2024ClothReview, KadiSensors2023CDOReview}. In this task, a robot grasps entangled cloth and progressively resolves the entanglement by adapting its pulling strategy \cite{HamajimaJRM1998LaundryUntangling,LonghiniArXiv2024ClothReview}. Cloth entanglement cannot be resolved by a single pulling motion and appears to require responses to contact and tension changes. Such adaptive behaviors are difficult to encode using hand-crafted strategies, and learning-based approaches are therefore more suitable. However, conducting large-scale real-world trials is impractical due to the risk of cloth damage and mechanical wear of the robotic hand \cite{DR-DRL-LSTM, legged_walking_sim2real, drl-door}. Therefore, this paper addresses robotic cloth untangling through a sim-to-real policy transfer framework \cite{rock_excavation_sim2real, pbrl_sim2real}, where policies are trained in simulation and directly deployed in the real world.

\begin{figure}[t]
    \hspace{-5mm}
    \centering
    \includegraphics[width=0.9\columnwidth]{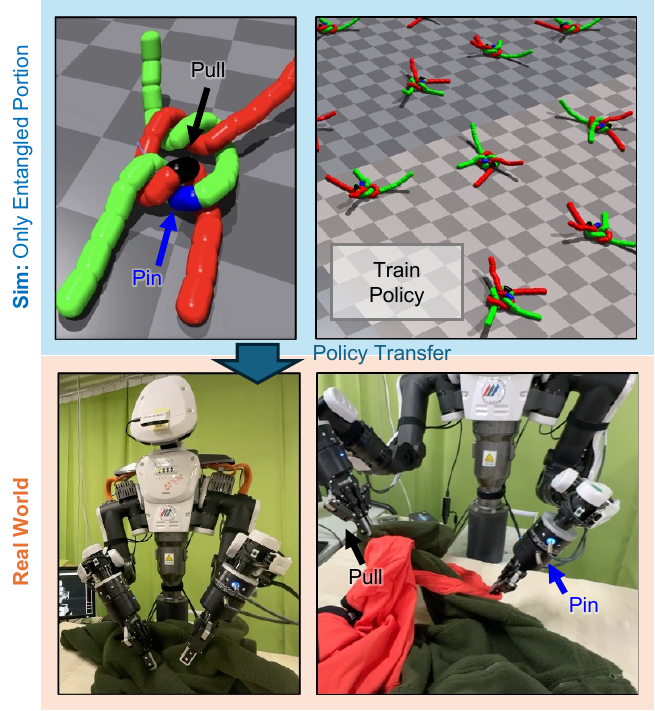}
    \caption{
        Overview of sim-to-real policy transfer for cloth untangling.
        In simulation, only the entangled portion of the cloth is modeled, and the policy is trained to progressively resolve the entanglement by fixing one hand (pin hand) and pulling with the other (pull hand). Then, the learned policy is transferred to the real world. 
    }
    \label{fig:task_overview}
\end{figure}

Cloth untangling is governed by contact interactions, friction, and tension variations, and even small parameter discrepancies between simulation and the real world can alter contact transitions and force responses \cite{DR-deformable,BlancoMuleroRAL2024Sim2RealGap}. Consequently, sim-to-real transfer is particularly challenging. Force information is essential for handling such interaction-dominant dynamics; however, discrepancies in contact and friction modeling often lead to mismatched force signals across domains. Policies that rely heavily on precise force magnitudes or small variations may therefore fail to obtain similar responses after transfer \cite{BlancoMuleroRAL2024Sim2RealGap,ZhangCoRL2023OnlineAdmittance}. Thus, a central challenge is to design observation representations that exploit force cues without excessive dependence on their absolute precision.

To address this issue, this paper revisits the objective of cloth untangling, which is to progressively resolve entanglement rather than to precisely regulate force magnitude \cite{HamajimaJRM1998LaundryUntangling,LonghiniArXiv2024ClothReview}. Although force information serves as a primary cue for action selection, what often matters for untangling is whether tension is increasing or decreasing, indicating contact transitions, rather than the exact force value \cite{bernard2025cooperative, sim2real_review}. We therefore hypothesize that emphasizing coarse force-change patterns while suppressing fine-grained, environment-dependent variations can better align with the task structure than directly minimizing the sim-to-real gap in force measurements. Concretely, we lower observation resolution in both simulation and the real world to mitigate overfitting to environment-specific force characteristics and facilitate direct sim-to-real transfer.

Based on this insight, this paper proposes Adaptive Force-Difference Quantization (ADQ), a policy learning framework that intentionally reduces observation resolution. ADQ represents force inputs using temporal differences instead of absolute values and discretizes them into three states: increase, decrease, or no change. This representation weakens dependence on absolute force scale and small variations while emphasizing contact transition tendencies shared across environments. Rather than fixing the discretization thresholds, ADQ treats them as policy-controlled variables, allowing the sensitivity of the representation to adapt to environmental conditions and task progression. Policies trained in simulation under this observation design are then deployed in the real world using the same representation, aiming to achieve robust behavior despite environment-dependent force variations.

To evaluate the proposed framework, we developed a simulation environment for policy learning and transferred the trained policies to the real world for cloth untangling experiments. We compared ADQ with policies that directly use raw force inputs. ADQ achieved higher task success rates in the real world while maintaining stable behavior across different friction properties and entanglement conditions. In contrast, policies using raw force inputs exhibited clear performance degradation after transfer, indicating poor robustness to environment-dependent force variations. These findings suggest that reducing observation resolution improves robustness of sim-to-real transfer compared to directly using raw force signals.

\section{Related work}
\label{s:related_works}

\subsection{Cloth Untangling and Deformable Object Manipulation}

    Cloth manipulation is a representative challenge in deformable object manipulation, characterized by high-dimensional configurations, self-occlusion, and contact-induced discontinuities. Early work addressed laundry separation and folding using heuristic planning and geometric reasoning \cite{HamajimaJRM1998LaundryUntangling, MillerIJRR2012LaundryFolding}. More recently, learning-based approaches integrating vision and tactile feedback have been proposed for complex cloth interactions \cite{LonghiniArXiv2024ClothReview,KadiSensors2023CDOReview}. In particular, disentanglement and dressing require long-horizon reasoning over successive contact transitions, and sim-to-real training has been explored in both dressing and untangling tasks \cite{ZhangSciRobotics2022Dressing,Sundaresan2021Untangling,Liu2023Robotic}.
    
    Compared to rope, cloth exhibits more severe occlusions and frictional variability, reducing the reliability of purely vision-based topology inference. Consequently, force feedback becomes essential for detecting interaction-state changes. However, deformable contact dynamics are difficult to reproduce accurately in simulation, leading to force discrepancies across domains. Robust sim-to-real transfer therefore requires force representations that remain informative under such mismatches.
    
    Unlike prior approaches that focus on improving simulation fidelity or directly using raw force signals, we emphasize observation design. Instead of matching absolute force distributions, we reshape force signals to highlight contact-state changes while suppressing environment-dependent variations. Cloth untangling is thus treated not as a problem of precise force reproduction, but as one of robust force representation under sim-to-real discrepancies.

\begin{figure}[t]
    \vspace{2.6mm}
    \centering
    \includegraphics[width=0.99\columnwidth]{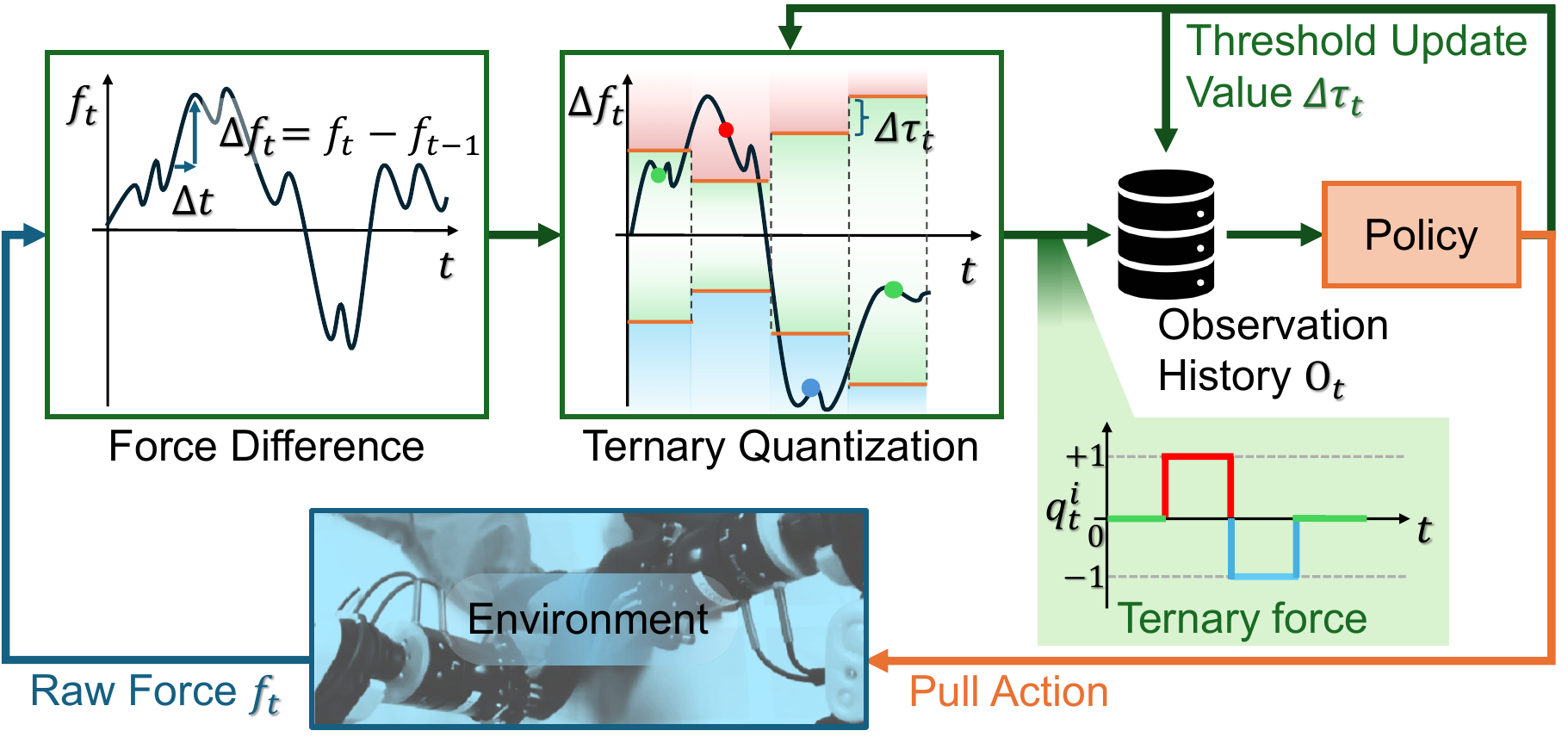}
    \caption{
        Overview of ADQ.
        Raw force measurements are converted into force differences and discretized into three states (increase: $+1$, decrease: $-1$, no change: $0$). The quantization thresholds are treated as policy-controlled variables, enabling adaptive sensitivity during execution. The resulting quantized force-difference representation is fed back into the policy to determine both the pulling actions and the thresholds.
        Policy training is performed within a reinforcement learning framework using this observation design.
    }
    \label{fig:method_overview}
\end{figure}

\subsection{Sim-to-Real for Contact-Rich Manipulation and Robust Contact-Signal Representations}

    Sim-to-real transfer in contact-rich manipulation has primarily focused on reducing discrepancies at the dynamics level. Common strategies include domain randomization~\cite{Tobin2017Domain}, active randomization~\cite{ActiveDR}, and parameter identification via optimization~\cite{simopt-baysian}, all of which aim to align simulated and real dynamics. Representation-level approaches have also been explored, for example through feature adaptation networks or real-to-sim signal translation to compensate for sensor differences~\cite{james2019sim,ChurchArXiv2021Real2SimTactile}. These methods typically attempt to preserve or recover the fidelity of continuous signals across domains.
    
    However, in contact-rich deformable manipulation, small mismatches in friction, compliance, or contact timing can induce structural distortions in force and tactile time series. In such settings, the issue is not only sensor bias or scale differences, but also domain-dependent reshaping of the signal trajectory itself~\cite{BlancoMuleroRAL2024Sim2RealGap,ZhangCoRL2023OnlineAdmittance}. When policies rely on precise continuous magnitudes, even well-randomized simulations may fail to reproduce the interaction patterns required for stable transfer.
    
    Unlike approaches that seek to better match or reconstruct raw contact signals, we adopt a different perspective: instead of increasing signal fidelity, we intentionally reduce observation resolution to suppress domain-specific variations. By emphasizing change-level interaction cues and discarding fine-grained magnitude information, our approach shifts the objective from force-distribution alignment to representation robustness. This positions our method as structurally distinct from prior dynamics-matching or signal-reconstruction strategies in contact-rich sim-to-real learning.

\section{Adaptive Force-Difference Quantization}

    This section presents ADQ, an observation design that mitigates sim-to-real discrepancies in force signals, which often differ across simulation and real robots (Fig.~\ref{fig:method_overview}). Because force measurements vary due to sensor scale, bias, noise, and contact stiffness, directly using raw continuous forces can lead to unstable transfer. ADQ addresses this issue through three components: (i) force differencing to emphasize change patterns, (ii) ternary quantization to provide robustness to distribution mismatch, and (iii) adaptive thresholding, where the policy controls quantization thresholds online. Together, these designs suppress environment-dependent variations while preserving contact transition cues relevant to the task.
    
\subsection{Force-Difference Representation}
    Let the 3-axis force signal at time $t$ be $f_t\in\mathbb{R}^3$. We focus on changes rather than absolute magnitudes and use
    \begin{equation}
    \Delta f_t = f_t - f_{t-1}
    \end{equation}
    Differencing relatively suppresses the effect of steady bias and emphasizes force rise/fall associated with switching contact states, yielding features that are more likely to be consistent between real and simulated environments.

\subsection{Ternary Quantization}
    For robustness to sim-to-real gaps (scale mismatch and noise mismatch), we coarsely quantize the difference signal $\Delta f_t$ into ternary values for each axis. For axis $i\in\{x,y,z\}$, using a threshold $\tau_t^i>0$, we define
    \begin{equation}
    q_t^i = Quant(\Delta f_t^i;\tau_t^i), \qquad
    Quant(x;\tau)=
    \begin{cases}
    -1 & x < -\tau \\
    0  & |x|\le\tau \\
    +1 & x > \tau
    \end{cases}
    \end{equation}
    and set $q_t=[q_t^x,q_t^y,q_t^z]^\top\in\{-1,0,+1\}^3$. Here, $q_t$ preserves the sign-level information of force changes---increase, no change, or decrease---while providing a discrete observation that is less sensitive to small noise and scale differences.

\subsection{Observation Design with Adaptation State}
    The observation $o_t$ concatenates the quantized force difference $q_t$, the previous threshold update $\Delta\tau_{t-1}\in\mathbb{R}^3$, and other observation states $z_t\in\mathbb{R}^n$ as $o_t = [q_t,\; \Delta\tau_{t-1}, z_t] \in \mathbb{R}^{6+n}.$
    By including $\Delta\tau_{t-1}$ in the input, the policy can estimate the recent quantization sensitivity (i.e., what magnitude of change is being detected as $\pm 1$), enabling stable adaptation based on the history of sensitivity adjustments. Since the exact computation of $\Delta f_t$ depends on the task control rate and action design, we estimate $\Delta f_t$ from the force sequence observed within each action interval in our implementation (Sec.~\ref{s:intra_action_df}).

\subsection{Action Design: Threshold-as-Action}
    The action $a_t$ concatenates the robot control input $u_t\in\mathbb{R}^m$ and the incremental update of the quantization thresholds $\Delta\tau_t\in\mathbb{R}^3$ as $a_t = [u_t,\; \Delta\tau_t].$ 
    Here, the pulling command $u_t \in [-1,1]^3$ represents a normalized Cartesian direction vector of the pull hand. 
    During execution, $u_t$ is scaled by a task-level parameter $\alpha>0$ specified in the experimental setup. In this work, we use  $\alpha = 3 \ {\rm{cm}}$.
    Each component of $u_t$ corresponds to the directional contribution along the $x,y,z$ axes, allowing the policy to modulate pulling direction in response to interaction-state changes.

    In this work, we use the same range constraint and initial value for all axes: $\tau_{\min}=0.005,\qquad \tau_{\max}=2.5,\qquad \tau_0=0.15$.
    The key idea is to treat the threshold $\tau$ not as a fixed preprocessing hyperparameter that must be tuned externally, but as a controllable variable embedded in the policy's action space via threshold-as-action.
    This enables the policy to autonomously adjust the resolution of force changes to be detected depending on the environment gap and contact conditions.

\subsection{Inference Procedure}
    At inference time, we first compute the force difference $\Delta f_t$ from the raw force signal $f_t$. 
    Since the ternary state depends on whether the change magnitude exceeds a threshold, the quantized observation is inherently determined by the current sensitivity parameter $\tau_t$. 
    We therefore apply ternary quantization using $\tau_t$ to obtain $q_t$, which encodes whether the recent interaction change is significant relative to the current sensitivity level.
    The resulting observation $o_t = [q_t, \Delta\tau_{t-1}, z_t]$ is fed into the policy $\pi_\theta$, which outputs both the pulling command $u_t$ and the threshold update $\Delta\tau_t$. 
    The threshold is then updated sequentially before the next timestep, allowing the policy to adjust its own observation sensitivity online according to the interaction phase and contact condition.

\section{Learning Cloth Untangling}
In this section, we describe the task formulation for learning the cloth untangling task within a sim-to-real framework, as well as the simulation environment we developed to support this learning process.

\subsection{Task Setting}
This work study garment-knot untangling as a controlled proxy of laundry disentanglement, focusing on local contact/tension transitions that dominate decision making.
Specifically, we consider a double-overhand knot formed using actual clothing such as sleeves and hems, which induces contact-rich transitions governed by friction and tension.
The robot grasps two garment points with two hands (pin/pull): The pin hand provides a fixed constraint, while the pull hand executes pulling motions.
We assume that a human provides a solvable grasp configuration.
However, the knot cannot be resolved by naively pulling left/right with brute force under a force budget, and thus the policy must switch strategies based on interaction cues.

We enforce a hard force limit such that the robot cannot pull beyond a threshold.
That is, the force is saturated/limited at execution time rather than penalized in the reward.
As a result, policies that simply repeat pulling actions are unlikely to succeed, and fine-grained switching of manipulation strategies conditioned on the contact state is required.

\subsection{POMDP Formulation and Temporal Observation Stack}
Since the full state of the cloth entanglement, such as knot topology and contact/friction conditions, is not directly observable, we formulate the task as a partially observable Markov decision process (POMDP).
Let the latent state at time $t$ be $s_t$, the observation be $o_t$, and the action be $a_t$.
The policy $\pi_\theta$ takes the observation sequence over the past $H$ steps as input and outputs an action.

\begin{figure}
    \centering
    \includegraphics[width=0.99\linewidth]{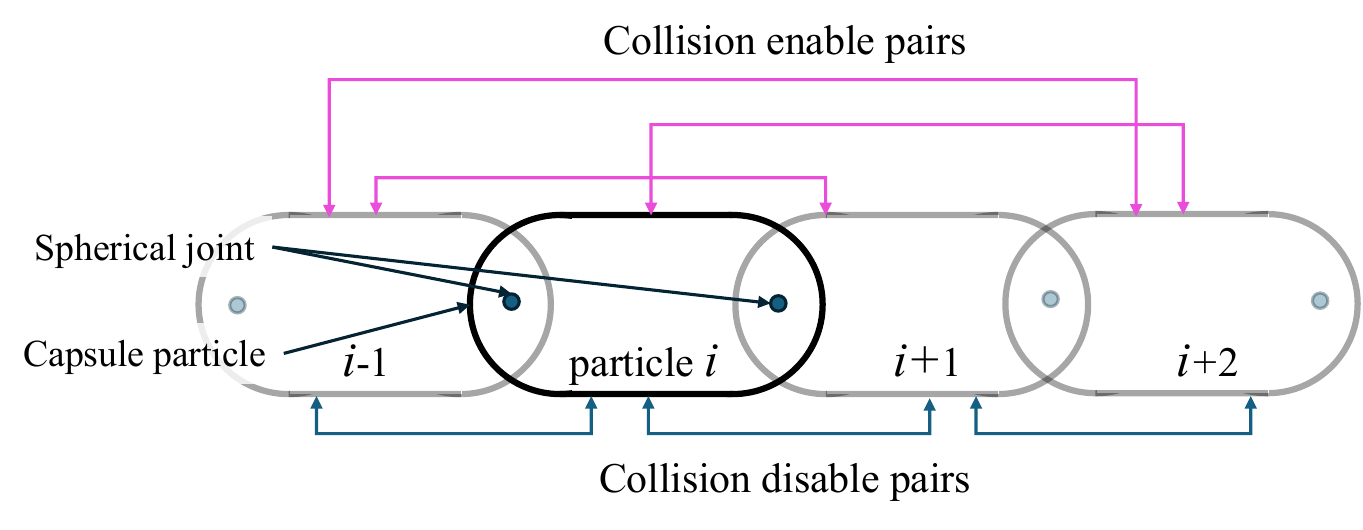}
    \caption{Local entanglement model for simulating cloth entangling. The entangled cloth portion is approximated as a chain of capsule segments connected by spherical joints where neighboring segments overlap to form a continuous body. To represent self-contact without introducing unstable persistent contacts, collisions are disabled for adjacent (overlapping) pairs and enabled only for non-neighboring pairs.}
    \label{fig:rope model stracture}
\end{figure}

\subsubsection{Intra-action force-difference estimation}
\label{s:intra_action_df}
In our task, one step (time $t$) corresponds to a single pulling motion executed by the pull hand.
Untying a double knot requires repeatedly switching subtle manipulations under a force limit. During a pull, friction/contact conditions can transition gradually, while instantaneous vibrations and noise can also be injected.
Therefore, rather than using a single difference at the step boundary, it is more appropriate to use a difference representation that stably summarizes the trend of changes within the action interval.

Specifically, let the force sequence observed at a high rate during the action execution interval at time $t$ be $\{f_{t,k}\in\mathbb{R}^3\}_{k=0}^{K}$, and define the intra-action difference sequence as
\begin{equation}
\delta f_{t,k} = f_{t,k} - f_{t,k-1}, \qquad k=1,\ldots,K.
\end{equation}
In this work, we use the average of these intra-action differences as the force difference observed in one step:
\begin{equation}
\Delta f_t = \frac{1}{K}\sum_{k=1}^{K}\delta f_{t,k}
          = \frac{1}{K}\sum_{k=1}^{K}(f_{t,k}-f_{t,k-1}).
\end{equation}
This averaging suppresses the effect of instantaneous fluctuations while preserving, as a step-level representation, the direction (increase/decrease) of contact transitions that occur during pulling.
The ternary quantization described below is applied to this $\Delta f_t$ to construct $q_t$.

\subsubsection{Temporal Observation Stack}
    We use a temporal stack of length $H=5$ and define the policy input as $\mathbf{O}_t = [o_{t-4},\ldots,o_t]$, where $a_t \sim \pi_\theta(\cdot \mid \mathbf{O}_t)$. 
    Each observation $o_t$ consists of the ternary-quantized force difference $q_t$, the most recent threshold update $\Delta\tau_{t-1}$, and a geometric feature $z_t$ defined as the direction vector from the pin hand to the pull hand. 
    Because contact states cannot be reliably inferred from a single timestep, stacking ${q_{t-H+1},\ldots,q_t}$ allows the policy to capture short-horizon change patterns such as sign flips and sustained trends that indicate interaction-state transitions.

\subsection{Reward and Success Criterion}
During training, we design the reward by combining (i) an entanglement metric $G_t$, (ii) the free-end length on the pulling side $\ell_t$, and (iii) a success indicator.
The entanglement metric $G_t$ is computed based on the Gauss linking integral \cite{Matsubara2013Reinforcement} and can be obtained from geometric information in simulation.
Let $\Delta\ell_t=\ell_t-\ell_{t-1}$ denote the increase in the free-end length. We define the reward as $r_t = w_{\ell}\,\Delta \ell_t - w_G\,G_t + w_s\,\mathbb{I}[\text{success}]$, 
where $w_{\ell}, w_G, w_s>0$ are weighting coefficients.
In this work, we regard the state where the free-end length on the pulling side disappears as the first loop of the double knot has been resolved, and define success as $\text{success} \Leftrightarrow \ell_t \le \varepsilon$
where $\varepsilon>0$ is a threshold.
Note that in real-robot experiments, $G_t$ cannot be computed directly. Therefore, we use $G_t$ for reward shaping during training, and evaluate performance in the real world using task-level metrics such as the success rate.

\subsection{Sim-to-Real Training Setup}

\subsubsection{Local Entanglement Model}
    Although training is performed in simulation, reproducing full cloth entanglement with high fidelity is challenging in terms of computational cost and numerical stability.
    Therefore, we adopt local entanglement model that captures the \emph{local} contact and tension transitions that dominate knot/entanglement resolution in cloth.
    Concretely shown in Fig.\ref{fig:rope model stracture}, we represent the interacting cloth part (e.g., sleeve/hem segment involved in the knot) as a chain of capsule-shaped segments with overlapping neighbors.
    This proxy preserves the essential interaction regime for force-driven decision making, while enabling stable large-scale RL training.

\subsubsection{Domain Randomization for Sim-to-Real}
    To facilitate sim-to-real transfer, we apply domain randomization during training.
    Specifically, we randomize, for each episode, (1) the scale and bias of the force-sensor signals, (2) the $z$-axis rotation of the object being pulled, and (3) the assignment of the pin/pull hand roles. In addition, we randomize (4) the hand orientation vector (observation error) at every step.
    This encourages learning policies that are robust to variations in observation distributions and contact transitions.


\begin{figure}[t]
    \centering
    \includegraphics[width=1.0\linewidth]{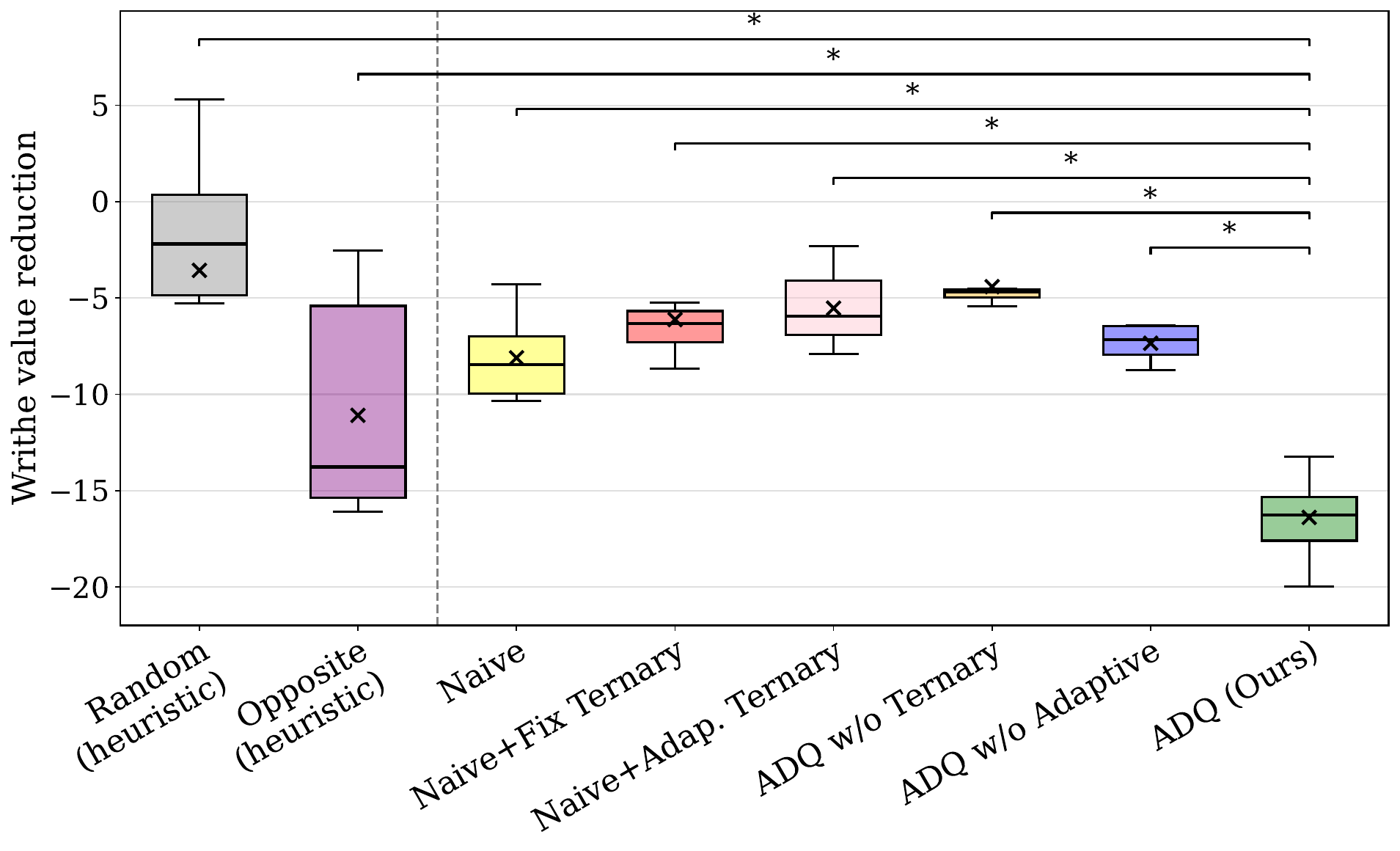}
    \caption{Sim-to-Sim performance comparison in Gazebo. Lower (more negative) writhe reduction indicates better untangling. Statistical significance is assessed using one-sided Welch's t-tests (Ours vs. each baseline; alternative: Ours is better), with Holm correction for multiple comparisons. * indicates significance level of $p<0.05$. Each method was evaluated over $n{=}10$ trials.}
    \label{fig:512}
\end{figure}

\begin{figure}[t]
    \centering
    \includegraphics[width=0.49\textwidth]{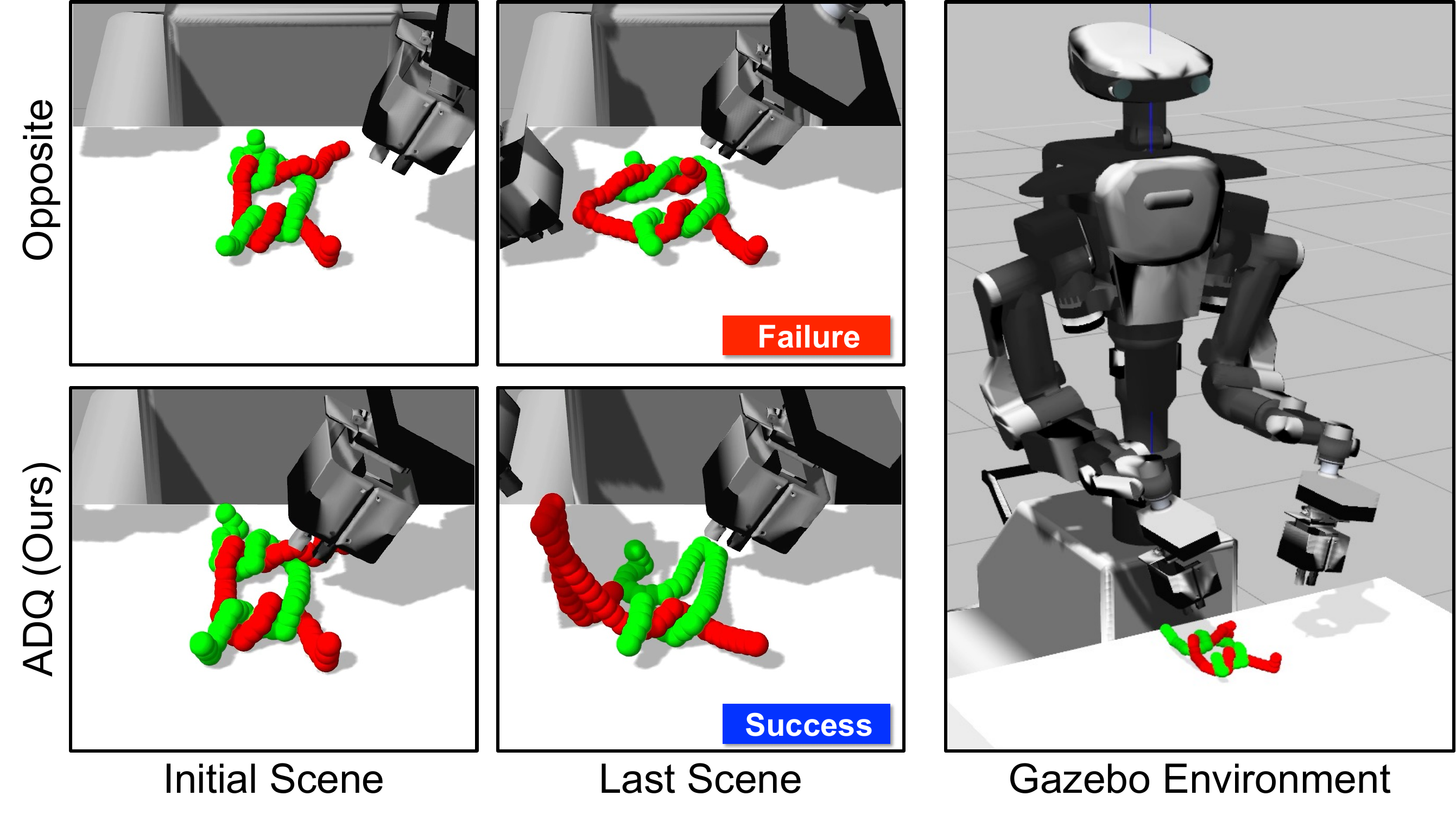}
    \caption{
        Snapshots of learned policy execution in the Gazebo simulator. 
        The initial scene before policy execution and the final scene after execution are shown. 
        An overview of the developed Gazebo-based cloth untangling environment is illustrated on the right side.
    }
    \label{fig:gazebo}
\end{figure}

\begin{figure}[t]
    \centering
    \includegraphics[width=0.99\linewidth]{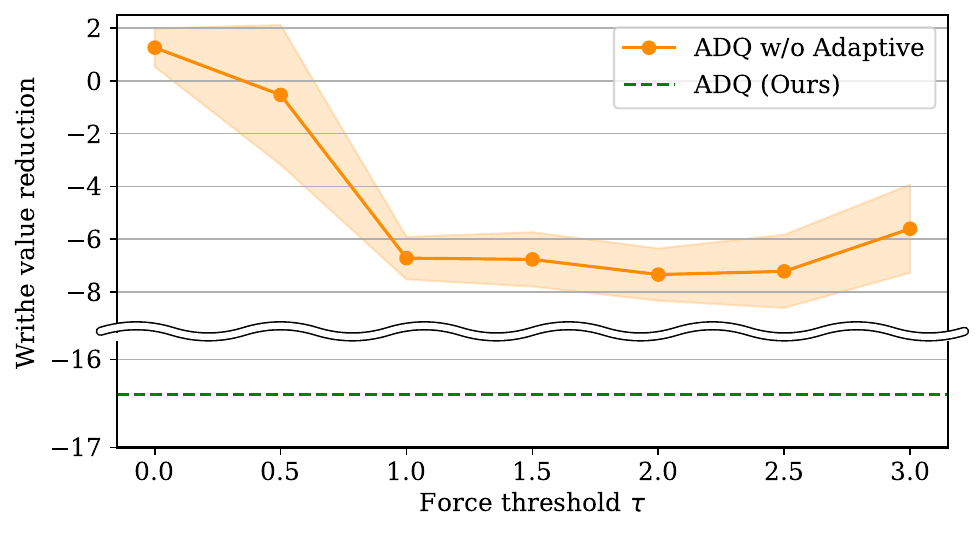}
    \caption{Threshold sensitivity in Gazebo. We sweep the fixed ternary quantization threshold $\tau$ for ADQ w/o Adaptive and compare it with ADQ (Ours), which adapts $\tau$ online. Although ADQ w/o Adaptive was trained with $\tau{=}0.5$ in Isaac Gym, its best-performing threshold in Gazebo shifts to $\tau{=}2.0$ (average writhe reduction $-7.3$), indicating a hyperparameter shift across simulators. In contrast, ADQ (Ours) remains stable and achieves larger writhe reduction, reducing the manual tuning burden.}
    \label{fig:513}
\end{figure}

\section{Experiments}
\label{sec:experiments}

    This section evaluates the proposed observation design from four perspectives: (i) whether the force-difference representation improves transfer robustness, (ii) whether coarse ternary quantization provides tolerance to scale and noise mismatch in force signals, (iii) whether adaptive thresholding reduces manual tuning across environments, and (iv) whether the learned policy relies on force features for action selection, analyzed using Grad-CAM~\cite{conf/iccv/SelvarajuCDVPB17}.
    They are validated through sim-to-sim transfer from Isaac Gym to Gazebo, where representation components can be systematically ablated and disentanglement progress quantified using simulator state. We then demonstrate end-to-end sim-to-real deployment on a physical robot with real cloth to verify transfer under sensing, execution, and material-dependent interaction gaps. 
    
    Policies are trained in Isaac Gym and evaluated in two stages: sim-to-sim transfer to Gazebo and sim-to-real deployment. Gazebo mirrors the real execution stack while providing access to the simulator state for quantitative analysis. Each subsequent subsection specifies its own setup and evaluation protocol.

\subsection{Learning Force-Based Policy in Simulation}
\label{sec:e1_learning}

Training was performed in the NVIDIA Isaac Gym framework on an NVIDIA GeForce RTX 4090 GPU.
The policy converged after approximately 157\,M training steps over a duration of 12 hours.
We set the policy's observable steps $H$ to 5 and apply domain randomization during training to facilitate sim-to-real transfer. Table~\ref{table:randomized-range} summarizes the randomized parameters and ranges. In addition, the initial local entanglement model orientation is randomized by sampling a yaw rotation about the world $z$-axis uniformly from $[0, 2\pi)$. We also randomize the grasp points from some handmade candidates.

\begin{table}[t]
    \vspace{1mm}
    \centering
    \caption{Ranges of randomized simulation parameters}
    \label{table:randomized-range}
    \setlength{\tabcolsep}{10pt} 
    {\footnotesize
    \begin{tabular}{l c c c}
        \toprule
        Randomized Variables & unit & min & max \\
        \midrule
        Force-sensor scale coefficient & [-]   & $0.95$ & $1.05$ \\
        Force-sensor bias              & N     & $-0.6$  & $0.6$  \\
        Pull-vector multiplier         & [-]   & $0.05$   & $10.0$ \\
        Linear density of each chain      & kg/m  & $0.267$  & $0.445$  \\
        Joint friction coefficient     & [-]   & $0.375$ & $0.625$ \\
        Joint damping coefficient      & [-]   & $0.375$ & $0.625$ \\
        Surface friction coefficient   & [-]   & $0.3$   & $0.5$  \\
        \bottomrule
    \end{tabular}
    }
\end{table}


\begin{figure}[t]
    \centering
    \includegraphics[width=0.95\linewidth]{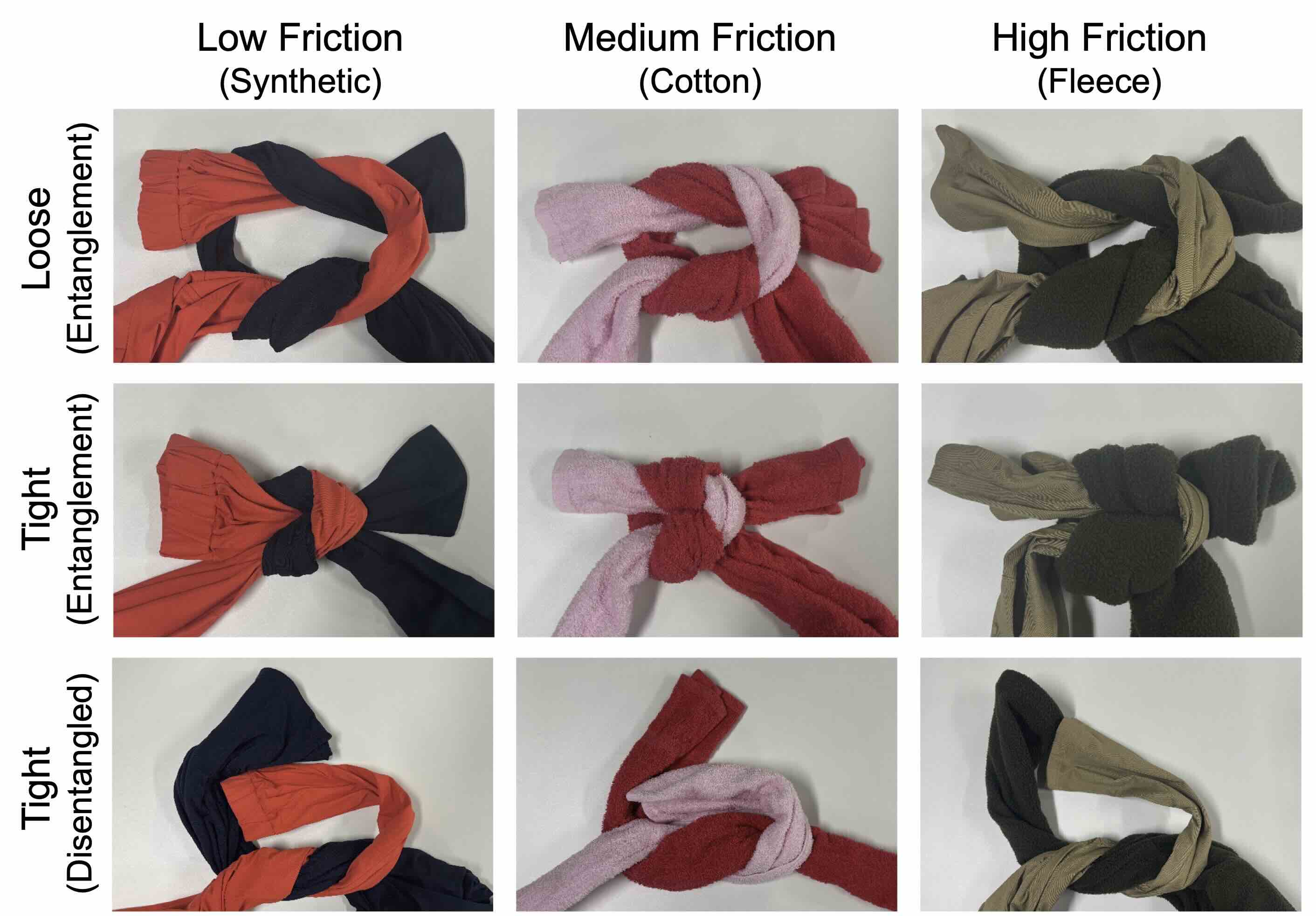}
    \caption{Experimental setup showing the cloth/textile configurations used during sim-to-real testing.}
    \label{fig:rope_configs}
\end{figure}

\begin{figure*}[t]
    \centering
    \includegraphics[width=0.99\textwidth]{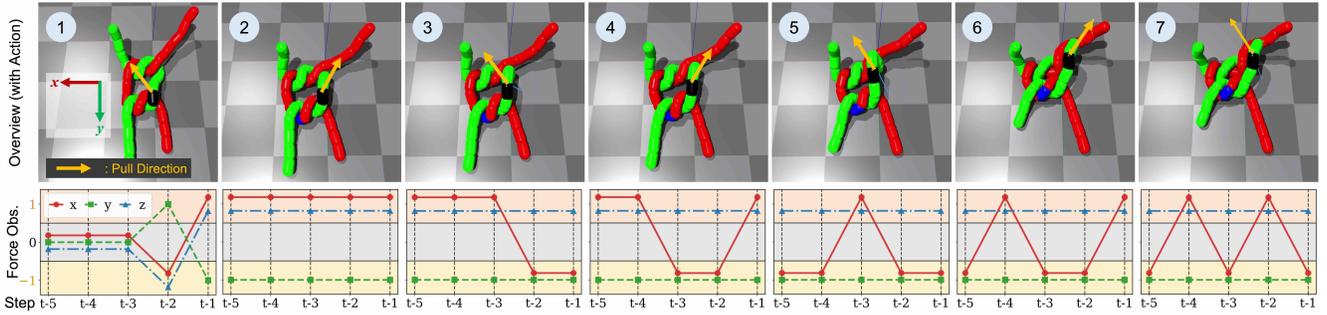}
    \caption{
        Snapshots of a tightly entangled case. 
        The commanded pulling force is overlaid as a yellow arrow in each frame. 
        The bottom plots show the observed force components $(x,y,z)$ over time. 
        Circled numbers denote the step index, and the step $t$ at the bottom corresponds to the highlighted frame.
    }
    \label{fig:514}
\end{figure*}

\subsection{Representation Ablation in Sim-to-Sim (Isaac Gym $\rightarrow$ Gazebo)}
\label{sec:e2_sim2sim_ablation}

This experiment analyzes how force differencing (FD), ternary quantization, and adaptive thresholding contribute to transfer robustness under a simulator gap. Policies trained in Isaac Gym are transferred to Gazebo, which mirrors the real execution stack (robot model, control interface, and safety limits) while isolating execution and contact-model discrepancies. Grasping is approximated using a rigid joint, and other settings follow the sim-to-real experiments.

We compare proposed \textbf{ADQ} against heuristic baselines (\textbf{Random}: a fixed random pulling direction per episode, and \textbf{Opposite}: pulling directly away from the pinning hand along the inter-hand line) and representation ablations/baselines derived from the learned policy. \textbf{Naive} removes both FD and ternary quantization, serving as a no-FD/no-quantization reference. To isolate the effect of ternary quantization without FD, we additionally evaluate \textbf{Naive+Fix Ternary} (ternary quantization with a fixed threshold) and \textbf{Naive+Adaptive Ternary} (ternary quantization with adaptive thresholding) while still omitting FD. We further test ablations of the proposed representation: \textbf{ADQ w/o Ternary} removes ternary quantization, and \textbf{ADQ w/o Adaptive} replaces adaptive thresholds with a fixed one. Disentanglement progress is quantified by a Gauss-integral-based entanglement metric computed from simulator state (writhe reduction), where more negative values indicate stronger disentanglement.

Fig.~\ref{fig:512} presents the performance comparison, and Fig.~\ref{fig:gazebo} shows representative snapshots of policy execution in Gazebo. 
\textbf{ADQ} achieves the largest writhe reduction, outperforming all baselines and ablations, which is consistent with the claim that emphasizing coarse force-change patterns improves transfer robustness. 
In contrast, variants that rely on raw force signals perform poorly. Adding ternary quantization alone does not improve \textbf{Naive}, and \textbf{Naive+Adaptive Ternary} yields the lowest performance, suggesting that raw force magnitudes are difficult to exploit reliably even with adaptive thresholding. 
Among FD-based variants, a clear progression is observed: \textbf{ADQ w/o Ternary} remains limited, \textbf{ADQ w/o Adaptive} provides moderate improvement, and the full \textbf{ADQ} achieves a substantial gain. These results indicate that robust transfer benefits from combining change-centered observations with coarse discretization and adaptive sensitivity to suppress environment-dependent force variations.


\subsection{Adaptive Thresholding Reduces Hyperparameter Tuning (Sim-to-Sim)}
\label{sec:e3_threshold_tuning}

This experiment evaluates the sensitivity of fixed-threshold ternary quantization (i.e., \textbf{ADQ w/o Adaptive}) and the stability of adaptive thresholding (i.e., \textbf{ADQ}).
A fixed quantization threshold can be optimal in the training simulator but suboptimal after transfer due to shifts in force scale and noise characteristics.
We focus on Gazebo evaluation and sweep the fixed quantization threshold over a range of values.
We compare this sweep to the proposed adaptive policy, which adapts the quantization threshold online during execution.
We use the same metric as in Sec.~\ref{sec:e2_sim2sim_ablation} (writhe reduction in Gazebo).
Fig.~\ref{fig:513} shows that the optimal fixed threshold shifts after transfer and performance becomes sensitive to $\tau$, whereas \textbf{ADQ} remains stable and achieves better disentanglement, reducing the need for manual tuning.


\subsection{Sim-to-Real Performance Across Clothing}
\label{sec:e4_sim2real}

We evaluated sim-to-real performance on the Nextage dual-arm humanoid robot equipped with a Robotiq FT 300 force-torque sensor and a 1-DoF gripper. To assess robustness to material-dependent interaction dynamics, experiments were conducted on clothing with different friction and stiffness (e.g., synthetic, cotton, thick textile), each tested in loose and tight double-overhand knot configurations. A trial was defined as successful if the knot was resolved within 15 pulls without exceeding a 30\,N force safety limit.
We compared \textbf{ADQ} with two baselines, \textbf{Opposite} and \textbf{Naive}. The success rates are summarized in Table~\ref{table:success-rates}. 
While the baselines performed adequately in low-friction and loose conditions, their performance degraded in high-friction or tight configurations that required force-aware adaptation. In particular, \textbf{Opposite}, which simply pulls in the opposite direction of the two end-effectors, often failed under high-friction conditions because such geometric heuristics cannot account for friction-induced resistance.
In contrast, \textbf{ADQ} achieved consistently high success rates across materials, indicating robustness to environment-dependent variations.
Efficiency and safety analysis further showed that although the baselines required fewer pulls in simple cases, \textbf{ADQ} maintained lower peak forces, reflecting stable closed-loop adaptation aligned with the proposed observation design.

\begin{table}[t]
    \vspace{1mm}
    \centering
    \caption{Success rate across materials and knot configurations under the sim-to-real protocol (success within 15 pulls and below a 30\,N force safety limit). Each configuration was evaluated over $n{=}10$ trials.}
    \label{table:success-rates}
    \setlength{\tabcolsep}{8pt} 
    {\footnotesize
    \begin{tabular}{llccc}
        \toprule
        \multicolumn{2}{l}{Configuration} & Opposite & Naive & ADQ (Ours) \\
        \midrule
        \multirow{3}{*}{\rotatebox{90}{Loose}} 
        & Low Friction    & 1.0 & 0.9 & 1.0 \\
        & Medium Friction & 1.0 & 0.9 & 1.0 \\
        & High Friction   & 0.9 & 0.6 & 0.9 \\
        \midrule
        \multirow{3}{*}{\rotatebox{90}{Tight}} 
        & Low Friction    & 0.9 & 0.5 & 0.9 \\
        & Medium Friction & 1.0 & 0.5 & 1.0 \\
        & High Friction   & 0.1 & 0.4 & 0.7 \\
        \midrule
        \multicolumn{2}{l}{Total Average} & 0.82 & 0.63 & \textbf{0.92} \\
        \bottomrule
    \end{tabular}
    }
\end{table}

\begin{table}[t]
    \vspace{1mm}
    \centering
    \caption{Efficiency and safety statistics over successful trials. We report the average number of pulls (pulls) and the peak force (peak), defined as the maximum force observed during a pulling action, for each configuration.}
    \label{table:pulls-and-forces}
    \setlength{\tabcolsep}{5pt} 
    {\footnotesize
    \begin{tabular}{ll ccc ccc}
        \toprule
        & & \multicolumn{2}{c}{Opposite} & \multicolumn{2}{c}{Naive} & \multicolumn{2}{c}{ADQ (Ours)} \\
        \cmidrule(lr){3-4} \cmidrule(lr){5-6} \cmidrule(lr){7-8}
        \multicolumn{2}{l}{Configuration} & pulls & peak & pulls & peak & pulls & peak \\
        \midrule
        \multirow{3}{*}{\rotatebox{90}{Loose}} 
        & Low Friction    & 6.9 & 9.3 & 7.3 & 8.1 & 9.7 & 7.1 \\
        & Medium Friction & 7.4 & 10.3 & 7.7 & 8.5 & 11.1 & 6.9 \\
        & High Friction   & 6.8 & 19.3 & 8.4 & 18.4 & 11.3 & 14.8 \\
        \midrule
        \multirow{3}{*}{\rotatebox{90}{Tight}} 
        & Low Friction    & 6.9 & 22.1 & 7.0 & 23.0 & 10.2 & 20.1 \\
        & Medium Friction & 7.0 & 23.0 & 7.6 & 19.8 & 9.3 & 17.2 \\
        & High Friction   & 6.0 & 23.4 & 9.0 & 28.5 & 8.7 & 19.2 \\
        \midrule
        \multicolumn{2}{l}{Total Average} & \textbf{6.8} & 17.9 & 7.8 & 17.7 & 10.1 & \textbf{14.2} \\
        \bottomrule
    \end{tabular}
    }
\end{table}


\begin{figure}[t]
    \centering
    \includegraphics[width=0.5\textwidth]{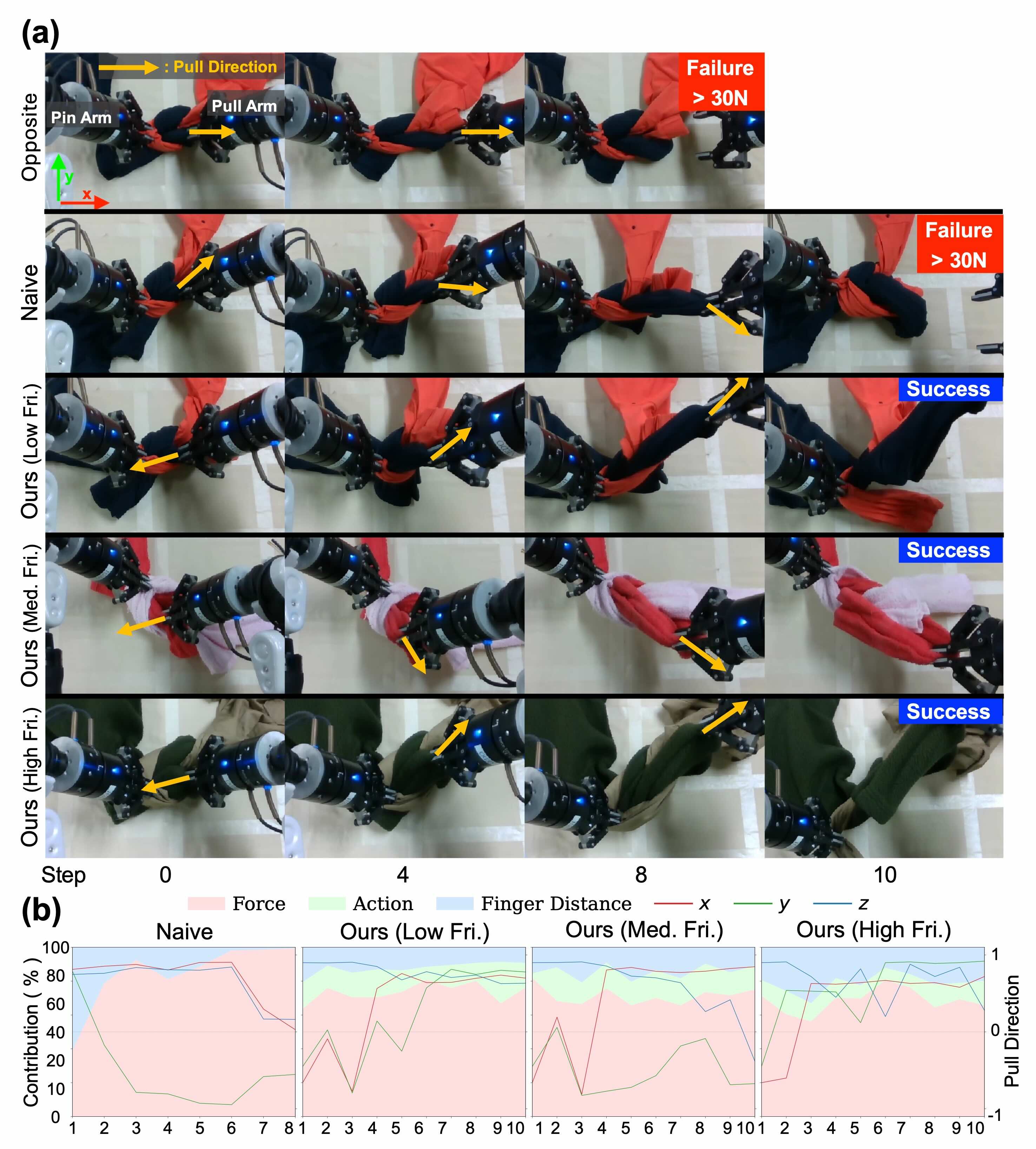}
    \caption{
        Analysis of policy behaviors. \textbf{(a)} Scene snapshots. The left arm is fixed, and the right arm pulls 3 cm per step. Yellow arrows denote the pulling direction. If the force exceeds 30 N, the gripper opens to ensure safety. The bottom three rows show different knot orientations and friction levels.
        \textbf{(b)} Grad-CAM features. Stacked areas indicate the feature contribution to the policy output. Solid lines show the pulling direction of $x, y$, and $z$.
        The pulling command $u_t$ is a normalized direction vector in $[-1,1]^3$.
        During execution, $u_t$ is scaled by $\alpha = 3 \ {\rm{cm}}$.
    }
    \label{fig:real_exp_all_figs}
\end{figure}

\begin{figure}[t]
    \centering
    \includegraphics[width=0.45\textwidth]{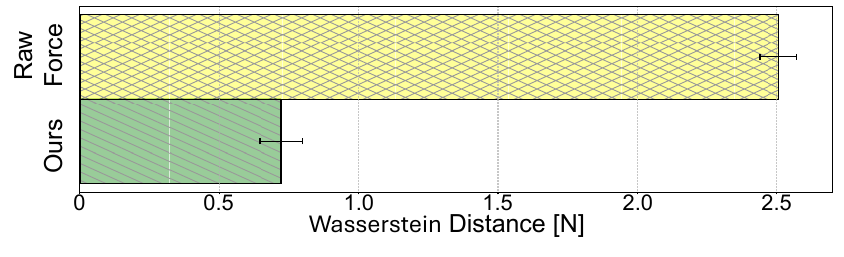}
    \caption{
        Sim-to-real gap measured by the 1D Wasserstein distance between observation distributions (Isaac Gym: $n{=}500$, Gazebo: $n{=}150$). Bars show bootstrap means with 95\% CI (2000 resamples).
    }
    \label{fig:real_exp_all_figs_c}
\end{figure}





\subsection{Explainability and Learned Strategy Analysis}
\label{sec:e5_explainability}

This experiment analyzes how the learned policy uses change-centered force features during execution and how adaptive thresholding maintains informative ternary-quantized FD signals.

\paragraph{Analysis of policy behavior in simulation (Fig.~\ref{fig:514})}
    Fig.~\ref{fig:514} illustrates representative rollout phases in simulation together with Grad-CAM attribution over time. The policy initially explores multiple pulling directions when interaction information is limited. During this early phase, attribution concentrates around recent force-difference features (e.g., near $k{-}1$), indicating that the policy uses short-window haptic cues to establish an initial pulling direction compatible with the current constraint.
    As execution proceeds, the policy maintains a dominant pulling direction while continuously monitoring the latest ternary-quantized FD features. When the sign of the $x$-component flips ($-1 \leftrightarrow +1$), attribution peaks around the change point and the policy switches its lateral pulling direction accordingly. This pattern suggests a closed-loop strategy: the policy detects discrete interaction-state changes through recent FD signals and updates the action to avoid worsening geometric constraints. Intervals where the $x$-component remains constant (e.g., $-1$) receive sustained attribution and correspond to consistent lateral pulling, while sign flips trigger directional reversal. These behaviors demonstrate that action updates are event-driven rather than dependent on absolute force magnitudes.

\paragraph{Analysis in the real world (Fig.~\ref{fig:real_exp_all_figs})}
    Fig.~\ref{fig:real_exp_all_figs}(a) shows snapshots of real-world execution, and Fig.~\ref{fig:real_exp_all_figs}(b) presents the corresponding Grad-CAM attribution. Despite sim-to-real discrepancies, ternary-quantized FD features consistently exhibit high attribution throughout the episode. Directional changes in the $x$-component of the pulling command align with sign transitions in the quantized FD signal, mirroring the simulation observations. This indicates that the learned strategy transfers: the policy continues to rely on recent force-difference events to detect contact or constraint transitions and adjust its pulling direction.
    Overall, the analysis supports the intended design principle of ADQ. The policy does not depend on precise force magnitudes; instead, it reacts to short-term, resolution-reduced force-difference signals to form a structured closed-loop disentanglement strategy.


\subsection{Sim-to-Real Gap in Force Representations}

To quantify whether the proposed representation mitigates the sim-to-real discrepancy in force observations, we execute the trained policy both in the training simulator (Isaac Gym) and on the real robot, and compare the empirical distributions of force-related observations across domains.
For each force axis, we compute the 1D Wasserstein distance $W_1$ between the two distributions as a measure of domain gap~\cite{Nicolas2017Optimal}. Fig.~\ref{fig:real_exp_all_figs_c} shows the comparison between raw force inputs and the proposed representation.
The baseline using raw force signals exhibits an average gap of $W_1 = 2.51$ (95\% CI: [2.44, 2.57]). In contrast, the proposed representation reduces the average gap to $0.71$ (95\% CI: [0.65, 0.80]), corresponding to an approximately 71\% reduction. This statistically significant reduction in $W_1$ indicates that the proposed observation design, based on force differences and ternary quantization, effectively reduces the sim-to-real gap in force-related observations.

\section{Discussion}
\label{s:dis}
Our results suggest that the bottleneck in leveraging force feedback for sim-to-real transfer lies not in reproducing the absolute magnitude of force, but in consistently capturing interaction changes induced by contact transitions.
Raw force signals can undergo substantial distribution shifts due to sensor bias, scale mismatches, and subtle differences in friction and grasp stiffness, which can cause the features relied on by the learned policy to break at transfer time.
In contrast, force differences (FD) are less sensitive to quasi-static offsets and slow drift. When combined with ternary quantization, they further gain invariance to noise and scale differences.
As a result, the domain gap in the observation space is reduced, making it easier for the closed-loop behavior to remain valid after transfer.

We discuss two limitations and future directions.
First, because our method discretizes force differences, performance may saturate on tasks where high-frequency vibrations or subtle force variations are critical.
Second, if the threshold updates are too rapid, the quantized representation can become non-stationary and destabilize learning. Therefore, it is important to design constraints and smoothing mechanisms (e.g., a penalty on the rate of change of the thresholds).
In future work, we will investigate multi-level quantization to gradually increase resolution, and evaluate generalization across a wider range of garment shapes, materials, and grasp conditions.
A natural direction for future work is to further isolate the effects of quantization and threshold adaptation on real clothing, where variability in fabric properties and knot tightness can be explicitly controlled.

\section{Conclusion}
\label{sec:conclusion}
We proposed an observation design for sim-to-real cloth untangling that reduces observation resolution by emphasizing coarse force-difference patterns rather than precise force magnitudes. The proposed policy learning framework ADQ suppresses environment-dependent force variations while preserving contact transition cues relevant to resolving entanglement. Policies trained in simulation achieved improved robustness and post-transfer performance in both Gazebo and the real world compared to raw force inputs, and feature attribution analysis confirmed reliance on quantized force-difference signals. These results demonstrate that task-aligned resolution reduction can enhance stability in sim-to-real transfer.
    

\bibliographystyle{IEEEtran}
\bibliography{reference}

@inproceedings{DR-DRL-LSTM,
  title={Sim-to-real transfer of robotic control with dynamics randomization},
  author={Peng, Xue Bin and Andrychowicz, Marcin and Zaremba, Wojciech and Abbeel, Pieter},
  booktitle={IEEE International Conference on Robotics and Automation},
  pages={3803--3810},
  year={2018},
}

@inproceedings{drl-door,
  title={Deep reinforcement learning for robotic manipulation with asynchronous off-policy  updates},
  author={Gu, Shixiang and Holly, Ethan and Lillicrap, Timothy and Levine, Sergey},
  booktitle={IEEE International Conference on Robotics and Automation},
  pages={3389--3396},
  year={2017},
}

@ARTICLE{Nicolas2017Optimal,
  author={Courty, Nicolas and Flamary, Rémi and others},
  journal={IEEE Trans. Pattern Anal. Mach. Intell.}, 
  title={Optimal Transport for Domain Adaptation}, 
  year={2017},
  volume={39},
  number={9},
  pages={1853-1865},
  doi={10.1109/TPAMI.2016.2615921}}

@article{Matsubara2013Reinforcement,
    author = {Takamitsu Matsubara and Daisuke Shinohara and others},
    title = {Reinforcement learning of a motor skill for wearing a {T}-shirt using topology coordinates},
    journal = {Advanced Robotics},
    volume = {27},
    number = {7},
    pages = {513--524},
    year = {2013}
}

@inproceedings{DR-deformable,
  title={Sim-to-real reinforcement learning for deformable object manipulation},
  author={Matas, Jan and James, Stephen and others},
  booktitle={Conf. Robot Learn.},
  pages={734--743},
  year={2018},
}

@inproceedings{james2019sim,
  title={Sim-to-real via sim-to-sim: Data-efficient robotic grasping via randomized-to-canonical adaptation networks},
  author={James, Stephen and Wohlhart, Paul and others},
  booktitle={IEEE/CVF Conf. Comput. Vis. Pattern Recognit.},
  pages={12627--12637},
  year={2019}
}

@article{simopt-baysian,
  title={Data-Efficient Domain Randomization With {B}ayesian Optimization},
  author={Muratore, Fabio and Eilers, Christian and others},
  journal={IEEE Robot. Autom. Lett.},
  volume={6},
  number={2},
  pages={911--918},
  year={2021},
}

@inproceedings{ActiveDR,
  title={Active domain randomization},
  author={Mehta, Bhairav and Diaz, Manfred and others},
  booktitle={Conf. Robot Learn. },
  pages={1162--1176},
  year={2020},
}

@inproceedings{Tobin2017Domain,
  title={Domain randomization for transferring deep neural networks from simulation to the real world},
  author={Tobin, Josh and Fong, Rachel and others},
  booktitle={IEEE/RSJ Int. Conf. Intell. Robots Syst.},
  pages={23--30},
  year={2017},
}

@ARTICLE{Liu2023Robotic,
  author={Liu, Fei and Su, Entong and others},
  journal={IEEE Robot. Autom. Lett.}, 
  title={Robotic Manipulation of Deformable Rope-Like Objects Using Differentiable Compliant Position-Based Dynamics}, 
  year={2023},
  volume={8},
  number={7},
  pages={3964-3971},
  doi={10.1109/LRA.2023.3264766}
}

@INPROCEEDINGS{Sundaresan2021Untangling, 
    AUTHOR    = {Sundaresan, Priya and Grannen, Jennifer and others}, 
    TITLE     = {{Untangling Dense Non-Planar Knots by Learning Manipulation Features and Recovery Policies}}, 
    BOOKTITLE = {Robot.: Sci. Syst.}, 
    YEAR      = {2021}, 
    ADDRESS   = {Virtual}, 
    DOI       = {10.15607/RSS.2021.XVII.013} 
}

@inproceedings{ZhangCoRL2023OnlineAdmittance,
  title={Efficient Sim-to-real Transfer of Contact-Rich Manipulation Skills with Online Admittance Residual Learning},
  author={Zhang, Xiang and Wang, Changhao and others},
  booktitle={Conf. Robot Learn. },
  pages={1621--1639},
  year={2023}
}

@inproceedings{ChurchArXiv2021Real2SimTactile,
  title={Tactile Sim-to-Real Policy Transfer via Real-to-Sim Image Translation},
  author={Church, Alex and Lloyd, John and others},
  booktitle={Conf. Robot Learn.},
  pages={1645--1654},
  year={2022}
}

@article{BlancoMuleroRAL2024Sim2RealGap,
  title        = {Benchmarking the Sim-to-Real Gap in Cloth Manipulation},
  author       = {Blanco-Mulero, David and Barbany, Oriol and others},
  journal      = {IEEE Robot. Autom. Lett.},
  volume       = {9},
  number       = {3},
  pages        = {2981--2988},
  year         = {2024},
  doi          = {10.1109/LRA.2024.3360814}
}

@article{LonghiniArXiv2024ClothReview,
  title        = {Unfolding the Literature: A Review of Robotic Cloth Manipulation},
  author       = {Longhini, Alberta and Wang, Yufei and others},
  journal      = {arXiv preprint arXiv:2407.01361},
  year         = {2024}
}

@article{KadiSensors2023CDOReview,
  title        = {Data-Driven Robotic Manipulation of Cloth-like Deformable Objects: The Present, Challenges and Future Prospects},
  author       = {Kadi, Halid Abdulrahim and Terzic, Kasim},
  journal      = {Sensors},
  volume       = {23},
  number       = {5},
  pages        = {2389},
  year         = {2023},
  doi          = {10.3390/s23052389}
}

@article{HamajimaJRM1998LaundryUntangling,
  title        = {Planning Strategy for Task Untangling Laundry - Isolating Clothes from a Washed Mass -},
  author       = {Hamajima, Kyoko and Kakikura, Masayoshi},
  journal      = {J. Robot. Mechatron.},
  volume       = {10},
  number       = {3},
  pages        = {244--251},
  year         = {1998},
  doi          = {10.20965/jrm.1998.p0244}
}

@article{MillerIJRR2012LaundryFolding,
  title        = {A Geometric Approach to Robotic Laundry Folding},
  author       = {Miller, Stephen and van den Berg, Jur and others},
  journal      = {Int. J. Robot. Res.},
  volume       = {31},
  number       = {2},
  pages        = {249--267},
  year         = {2012},
  doi          = {10.1177/0278364911430417}
}

@article{ZhangSciRobotics2022Dressing,
  title        = {Learning garment manipulation policies toward robot-assisted dressing},
  author       = {Zhang, Fan and Demiris, Yiannis},
  journal      = {Sci. Robot.},
  volume       = {7},
  number       = {65},
  pages        = {eabm6010},
  year         = {2022},
}

@inproceedings{conf/iccv/SelvarajuCDVPB17,
  author = {Selvaraju, Ramprasaath R. and Cogswell, Michael and others},
  booktitle = {IEEE/CVF Int. Conf. Comput. Vis.},
  pages = {618-626},
  title = {Grad-CAM: Visual Explanations from Deep Networks via Gradient-Based Localization.},
  year = {2017}
}

@inproceedings{bernard2025cooperative,
  title={Cooperative grasping and transportation using multi-agent reinforcement learning with ternary force representation},
  author={Bernard-Tiong, Sheng and Tsurumine, Yoshihisa and Sota, Ryosuke and Shibata, Kazuki and Matsubara, Takamitsu},
  booktitle={IEEE/SICE International Symposium on System Integration},
  pages={973--978},
  year={2025},
}

@article{rock_excavation_sim2real,
  title={Progressive-resolution policy distillation: Leveraging coarse-resolution simulations for time-efficient fine-resolution policy learning},
  author={Kadokawa, Yuki and Tahara, Hirotaka and Matsubara, Takamitsu},
  journal={IEEE Transactions on Automation Science and Engineering},
  year={2025},
  volume={22},
  pages={18682-18693}
}

@article{pbrl_sim2real,
  title={DAPPER: Discriminability-Aware Policy-to-Policy Preference-Based Reinforcement Learning for Query-Efficient Robot Skill Acquisition},
  author={Kadokawa, Yuki and Frey, Jonas and Miki, Takahiro and Matsubara, Takamitsu and Hutter, Marco},
  journal={IEEE Robotics \& Automation Magazine},
  year={2026},
  pages={2-17},
}

@article{sim2real_review,
  title={A review of physics simulators for robotic applications},
  author={Collins, Jack and Chand, Shelvin and others},
  journal={IEEE Access},
  volume={9},
  pages={51416--51431},
  year={2021},
  publisher={IEEE}
}

@INPROCEEDINGS{legged_walking_sim2real,
  author={Watanabe, Ryo and Miki, Takahiro and Shi, Fan and Kadokawa, Yuki and Bjelonic, Filip and Kawaharazuka, Kento and Cramariuc, Andrei and Hutter, Marco},
  booktitle={IEEE International Conference on Robotics and Automation}, 
  title={Learning Quiet Walking for a Small Home Robot}, 
  year={2025},
  volume={},
  number={},
  pages={15285-15291}
}

\end{document}